# L1 Augmented Attention as an Improved Vector Similarity Metric

Kurt Godden
Walsh College, Troy Michigan, USA
kgodden@walshcollege.edu

Abstract:
Scaled dot-product attention conflates directional alignment and vector magnitude, limiting its effectiveness as a similarity metric in Transformer models. We introduce *L1-augmented attention*, a simple and computationally parallelizable modification that subtracts a learned, head-specific L1 distance between queries and keys from the dot-product score. This hybrid similarity captures complementary geometric information: dot-product rewards directional alignment, while L1 penalizes coordinate-wise deviations. To reduce the cost of L1 computation, we project queries and keys into low-dimensional subspaces whose parameters specialize to preserve informative L1 structure. Evaluated on WikiText-2 using a compact Transformer, L1-augmented attention achieves up to a 14.5% reduction in perplexity over the Vaswani baseline and outperforms an RBF-L2 kernel. Analysis of norm variance and learned L1 weights reveals distinct geometric roles across layers and strong head-level specialization. These results demonstrate that enriching attention with L1 geometry provides a principled and effective improvement to similarity computation in modern language models, with practical benefits for both accuracy and parallel efficiency.

## 1. Introduction

Note: This is a preliminary report. Additional experiments, variance analysis, and reproducibility enhancements will be included in a future revision.

The transformer architecture (Vaswani et al., 2017) crucially relies on scaled dot product attention, defined as:

(1) $$Attention(Q, K, V) = softmax\left(\frac{QK^\top}{\sqrt{d_k}}\right)V$$

The implicit assumption in (1) is that attention weights arising from the dot products of queries with keys measure similarity of the query and key vectors (Li et al., 2019), indirectly capturing related semantics between input tokens. Both queries and keys are derived from a lexicon of word or token embeddings, which themselves are usually learned concurrently with their associated query and key vectors, either in an autoregressive language model or a task-specific model such as text classification.

For a language model, the role of the query vector is to seek key vectors in the same text (or utterance in spoken language) that are important, to a greater or lesser extent, in predicting the next token. In the words of Clark et al. (2019), “Attention weights can be viewed as governing how ‘important’ every other token is when producing the next representation for the current token.”

In equation (1), the larger dot product values result in larger softmax probabilities with participating keys therefore treated as more important for predicting the next token.

Notice that the assumption of probability weights determining token importance encompasses an inductive bias that dot product is a *reliable* measure of similarity between a query vector $q_i$ and a key vector $k_j$.

In general, the phenomenal success of transformer based large language models validates this bias. However, there is also a problem that arises from the bias because the weights are ultimately derived from the dot product values, which in isolation, hide important angle and vector magnitude information that is implicit in dot product:

(2) $q_i \cdot k_j = \| q_i \| \| k_j \| cos(\theta)$, for the angle $\theta$ between $q_i$ and $k_j$.

Since the dot product value is a scalar, it masks the norms of the vectors to which it is sensitive (Kobayashi, et al., 2020). While we will unmask the norms in section 3, let us now illustrate not only successful, but also unsuccessful uses of dot product as a metric for vector similarity. For purposes of discussion, let us regard the L1 norm of vector difference as a baseline metric for dissimilarity between vectors. If we take the difference of any two identical vectors, the result is the zero vector, whose L1 norm is zero, which is to say there is no dissimilarity for identical vectors.

The absolute value of vector difference allows us to decompose the two vectors' dissimilarity into each dimension's contribution. Any dimension with an absolute difference greater than zero is that dimension's contribution to the overall dissimilarity. With this view of vector dissimilarity, let us investigate how reliably dot product alone can determine vector similarity. For our examples, we use the following three queries and five keys:

$$q_1 = \begin{pmatrix} 0.2 \\ 0.3 \\ 0.4 \end{pmatrix} q_2 = \begin{pmatrix} 0.3 \\ 0.4 \\ 0.5 \end{pmatrix}$$

$$k_1 = \begin{pmatrix} 0.9 \\ 0.8 \\ 0.7 \end{pmatrix} k_2 = \begin{pmatrix} 0.8 \\ 0.7 \\ 0.6 \end{pmatrix} k_3 = \begin{pmatrix} 0.9 \\ 0.8 \\ -0.7 \end{pmatrix} k_4 = \begin{pmatrix} 1.8 \\ 1.2 \\ 0.9 \end{pmatrix}$$ as well as

$$q_3 = k_5 = \begin{pmatrix} 0.6 \\ 0.6 \\ 0.6 \end{pmatrix}$$

With these vectors, if dot product were a reliable metric, we would expect the relation of dot products ($a_{dp}$): $q_1 \cdot k_1 < q_2 \cdot k_2$ to hold because each of the paired dimensions in $q_2$ and $k_2$ have a smaller difference (e.g. 0.3 and 0.8) than the corresponding pairs in $q_1$ and $k_1$ (0.2 and 0.9). According to our reference metric, $q_2$ is more similar to $k_2$ than $q_1$ is to $k_1$. Computing the dot products confirms our expectation:
$q_1 \cdot k_1 (= 0.7) < q_2 \cdot k_2 (= 0.82)$

Consider ($b_{dp}$): $q_1 \cdot k_1 > q_1 \cdot k_3$. The only difference in these two pairs of vectors lies in the two keys, $k_1$ and $k_3$, which differ only in the sign of their third dimension (0.7 vs. -0.7). That difference leads us to expect the relation to be true, which it is:
$q_1 \cdot k_1 (= 0.7) > q_1 \cdot k_3 (= 0.14)$

The inductive bias is again validated. When matched coordinates have opposite signs their product is subtracted during the dot product summation, reducing the dot product score. Dot product rewards copacetic directionality per dimension. Thus implicitly, dot product is still responsive to vector directionality even though from the dot product value we cannot explicitly identify the directionality captured by $cos(\theta)$.

However, dot-product conflates directional alignment with raw coordinate magnitudes. Large coordinates in a key vector can disproportionately increase the dot product, despite being less structurally similar to the query. (Wang, et al., 2020)

For instance, if dot product were an infallible similarity metric, then we would expect ($c_{dp}$): $q_1 \cdot k_1 > q_1 \cdot k_4$ to be true because the dimensional values in $k_4$ are all greater than the corresponding values in $k_1$ and therefore less similar to $q_1$ than $k_1$ is by our L1 norm metric. However, this time the relation is false:
$q_1 \cdot k_1 (= 0.7) > q_1 \cdot k_4 (= 1.08)$

Even worse is the following example. Since $q_3$ and $k_5$ are identical vectors, the L1 norm of their difference is zero, indicating perfect similarity. But comparing those identical vectors with dissimilar vectors, ($d_{dp}$) $q_3 \cdot k_5 > q_1 \cdot k_4$ is not true.

Rather, it is true that $q_3 \cdot k_5 = q_1 \cdot k_4 = 1.08$ since the pairwise multiplications in both $q_3 \cdot k_5$ and $q_1 \cdot k_4$ all equal 0.36 for every dimension. For example, the product of the first paired dimension in $q_3 \cdot k_5$ is $0.6 * 0.6 = 0.36$, just as in the first paired dimension from $q_1 \cdot k_4$ where $0.2 * 1.8 = 0.36$.

Thus, in this case dot product fails the similarity test, because it views the two vector pairs as equally similar despite one pair being identical and the other pair being dissimilar.

We should expect more from a similarity metric than what we obtain with dot product alone. This is not to claim that the L1 norm should replace dot product in the attention equation because L1, if used in isolation, also has its shortcomings. For example, while dot product is blind to the vector norms, the L1 norm is blind to dimensional directionality since the difference vector is converted to absolute values before summation.

In section 2 we explore some previous efforts to improve the attention equation of (1), and in section 3 we show how augmenting dot product with an L1 term leads to much improved similarity computation.

## 2. Related Work

A wide variety of modifications to the attention equation have been proposed to address different issues, but few explicitly address the conflation of magnitude and direction. For example, Shaw et al. (2018) modified attention to incorporate relative distance between a query and its associated keys.  Huang et al. (2019) made subsequent revisions to this idea in order to reduce memory requirements. The relative distance attention in Huang adds a term to supplement the dot product by referencing a matrix of learned embeddings, one for each relative distance between a query token and the keys that it attends to.

(3) $$Relative\,Attention = softmax\left(\frac{QK^\top + S^{rel}}{\sqrt{d_k}}\right)V$$

Shen et al. (2018) extend self-attention to create attention scores for every feature of the key vectors, rather than the entire token as a whole, allowing them to identify which features of a key are important to the query in context. Their primary goal is to create an embedding for the entire input sequence.

Guo et al. (2024) introduce a new process to identify tokens to prune from the key-value cache in order to economize memory for large language models that have extremely large context windows. Previous work generally focused on small attention scores to identify tokens that can be pruned with minimal impact on the attention head’s output.

The innovation by Guo et al. is to consider the L1 norm of value vectors in computing token importance. They observe that value vectors which correspond to large dot product attention scores tend to have small L1 norms. Thus, they use the following equation to compute the importance of token k at decoding step t.

(4) $$I_k^t = S_k^t \cdot \|v_k\|_1$$

where the first term on the right is the attention score and the second is the L1 norm of the value vector. They also include special treatment of attention sink tokens, such as start and end of sentences, so as not to delete them, given that their value norms are quite close to zero in almost all heads. Although they reference the L1 norm, the norm is not used to compute the similarity of query and key vectors. Just as Guo et al. compute the norms of value vectors, so do Kobayashi et al. (2020), although they use an RBF kernel with L2 norms.

More closely related to the work presented here are various other applications of RBF kernels with L2 norms in the attention equation. For example, Kim et al. (2021) substitute an RBF kernel with L2 attention in place of the standard dot product attention. Using individual queries and keys instead of matrices, their equation for the softmax score would be:

$$(5)\quad P_{ij} = softmax_j\left(-\frac{\|q_i - k_j\|_2^2}{2\sqrt{d}_k}\right)$$

However, their primary motivation is Lipschitz continuity, not similarity of vector geometry. A Lipschitz function is such that if you change the function's input by amount $\delta$, then the function's output cannot change by more than $K\delta$ for some constant K. In other words the function's derivative is bounded, which becomes very useful in some applications. As the authors write, "Lipschitz constraints can endow models with provable robustness against adversarial perturbations …, and guaranteed generalisation bounds."

Even though Kim et al. do not directly address dot product's problems, this RBF kernel is similar to what we propose below and we include a comparison of our method with theirs. The RBF-L2 kernel is also addressed in Tsai et al. (2019) in a broader discussion of using different kernels for attention. They compare the performance of several different kernels, including scaled and softmaxed dot product, RBF, polynomial and even linear. After testing different variants they conclude that RBF performs best for neural machine translation while exponential scaled and softmaxed dot product gives the best results for sequence predictions, i.e. language models.

But just as we saw for Lipschitz conformance, the primary concern in Tsai et al. is not vector similarity. Their interest is in how different kernels perform in general, and they reinterpret attention as kernel regression. Our L1-augmented similarity can be viewed as defining a new kernel that combines directional alignment via dot product with coordinate-wise geometric differences from the L1 term, cf. section 7 below.

One particular topic of interest to Tsai et al. is how the choice of kernel can lead to a "better way to integrate the positional embedding." This latter notion occupies its own subsection in their paper, as well as a prominent mention in the paper's abstract.

One result that is both relevant and important to the present work is that of Katharopoulos et al. (2020). They discuss a generalization of the Vaswani attention equation that substitutes a similarity function $sim(q, k)$ that is constrained to produce non-negative values. They further revise this generalized attention equation to take advantage of the associative property of matrix multiplication which results in an attention equation that scales linearly, rather than quadratically as in the softmax-based attention. We will return to their idea briefly in section 7.

We now turn to the topic of the present study, which focuses on reducing the failings of dot product as a metric for vector similarity by combining dot product with the L1 norm.

## 3. L1 Augmented Attention

To address the limitations of dot product as a vector similarity metric, we propose augmenting the scaled dot product with an L1 decrement:

(6) L1 Attention$(Q, K, V, H) = \text{softmax}\left(\frac{QK^\top}{\sqrt{d_k}} - \lambda_h [\|q_i - k_j\|_1]_{i,j}\right) V,$

where H refers to the attention heads to which the weighting parameter $\lambda_h$ is indexed, $Q, K \,\epsilon\, \mathbb{R}^{Txd_k}$ and $[\|q_i - k_j\|_1]_{i,j} \,\epsilon\, \mathbb{R}^{TxT}$ is the matrix of pairwise L1 distances between queries and keys. As we will see, this explicitly unmasks the otherwise hidden norm information and provides access to a richer geometry when calculating vector similarity.

For the moment, if we ignore $\sqrt{d_k}$ and assume that the L1 weighting factor $\lambda_h = 1$, let us see what the attention scores for the vectors of section 1 above would become. Here again are the four previous dot product relations with their values.

($a_{dp}$): $q_1 \cdot k_1 < q_2 \cdot k_2 = 0.70 < 0.82$ Satisfied
($b_{dp}$): $q_1 \cdot k_1 > q_1 \cdot k_3 = 0.70 > 0.14$ Satisfied
($c_{dp}$): $q_1 \cdot k_1 > q_1 \cdot k_4 = 0.70 > 1.08$ Not Satisfied
($d_{dp}$): $q_3 \cdot k_5 > q_1 \cdot k_4 = 1.08 > 1.08$ Not Satisfied

Let us now see how those same relations compare when the L1 term is added.

$$q_1 = \begin{pmatrix} 0.2 \\ 0.3 \\ 0.4 \end{pmatrix}, k_1 = \begin{pmatrix} 0.9 \\ 0.8 \\ 0.7 \end{pmatrix} \Rightarrow q_1 \cdot k_1 - \|q_1 - k_1\|_1 = 0.7 - 1.5 = -0.8$$

$$q_2 = \begin{pmatrix} 0.3 \\ 0.4 \\ 0.5 \end{pmatrix}, k_2 = \begin{pmatrix} 0.8 \\ 0.7 \\ 0.6 \end{pmatrix} \Rightarrow q_2 \cdot k_2 - \|q_2 - k_2\|_1 = 0.82 - 0.9 = -0.08$$

($a_{L1}$): $q_1 \cdot k_1 - \|q_1 - k_1\|_1 < q_2 \cdot k_2 - \|q_2 - k_2\|_1 = -0.8 < -0.08$ Satisfied

Both the decremented and non-decremented versions of relation (a) are treated as desired.

Similarly, the decremented and non-decremented versions of relation (b) are treated correctly because

$$q_1 = \begin{pmatrix} 0.2 \\ 0.3 \\ 0.4 \end{pmatrix}, k_3 = \begin{pmatrix} 0.9 \\ 0.8 \\ -0.7 \end{pmatrix} \Rightarrow \; q_1 \cdot k_3 - \|q_1 - k_3\|_1 = -2.16$$

Given this result and the one we just saw for $q_1 \cdot k_1 - \|q_1 - k_1\|_1$ we have:

($b_{L1}$): $q_1 \cdot k_1 - \|q_1 - k_1\|_1 > \; q_1 \cdot k_3 - \|q_1 - k_3\|_1 = -0.8 > -2.16$ Satisfied

And with our L1 modification, we now see that the anomalies above involving $q_1 \cdot k_4$ vis-a-vis $q_3 \cdot k_5$ are avoided:

$$q_1 = \begin{pmatrix} 0.2 \\ 0.3 \\ 0.4 \end{pmatrix}, k_4 = \begin{pmatrix} 1.8 \\ 1.2 \\ 0.9 \end{pmatrix} \Rightarrow \; q_1 \cdot k_4 - \|q_1 - k_4\|_1 = -1.92$$

instead of the unpenalized $q_1 \cdot k_4 = 1.08$. The penalized version here is now smaller than the penalized form for $q_1$ and $k_1$ (–0.8), giving us:

($c_{L1}$): $q_1 \cdot k_1 - \|q_1 - k_1\|_1 > \; q_1 \cdot k_4 - \|q_1 - k_4\|_1 = -0.8 > -1.92$ Satisfied

And finally, the absurdity above with identical vectors returning the same similarity value as quite different vectors is now resolved:

$q_3 = k_5 = \begin{pmatrix} 0.6 \\ 0.6 \\ 0.6 \end{pmatrix} \Rightarrow \; q_3 \cdot k_5 - \|q_3 - k_5\|_1 = 1.08$ with its zero decrement compared to $q_1 \cdot k_4 - \|q_1 - k_4\|_1$ that we just computed above, gives us:

($d_{L1}$): $q_3 \cdot k_5 - \|q_3 - k_5\|_1 > \; q_1 \cdot k_4 - \|q_1 - k_4\|_1 = 1.08 > -1.92$ Satisfied

The source of the L1 improvement is that L1 contains its own complementary inductive bias. Whereas dot product rewards vectors with similar directionality and sufficiently similar magnitudes, L1 directly measures coordinate-wise differences, penalizing deviations linearly. This preserves the proportional geometry of each dimension, enabling the model to represent finer-grained geometric distinctions between query and key vectors. In this sense, L1 attention increases the *expressivity* of the similarity metric, allowing the model to fit the geometry of token embeddings more precisely than either the dot product or the L1 norm alone.

All of the examples presented above use 'toy' vectors for clarity, and were deliberately constructed to highlight the strengths and weaknesses of dot product as well as L1 norms. However, we need to perform an unbiased test of the L1 augmentation in equation (6) using real data before we can claim that it is a serious improvement over the baseline attention of Vaswani et al. Thus, we next describe some implementation details of L1 augmentation, followed in section 5 by a discussion and performance comparison of L1 with the baseline dot product in (1) as well as RBF with L2 in (5).

# 4. Implementation

The Vaswani attention equation is implemented by Keras/TensorFlow in the MultiHeadAttention class. We can modify the equation by subclassing that parent as ModifiedMultiHeadAttention, copying the parent code for equation (1), and augmenting it to calculate the $\lambda$-weighted L1 term of equation (6). We then subtract the L1 norm from the scaled dot-product score and pass the result to softmax. A three-way software toggle controls whether the Vaswani baseline, the RBF L2, or the L1 code is executed, allowing us to train comparison models for all three. In order to maintain valid comparisons, the code was seeded to deterministically control random processes such as parameter initializations.

The original transformer architecture was not precisely duplicated due to constraints of available computational power. Instead of six transformer blocks with eight attention heads each and a model dimensionality of 512, two blocks of three heads each with a model dimensionality of 192 was implemented. Thus, each of the attention heads worked with the same key dimensionality (64) as in Vaswani.

The queries and keys are projected to a lower dimension (c.f. section 5) before computing the L1 norms. These projections are implemented using linear (Dense) layers with Glorot uniform initialization and no bias. Although the projection matrices are shared across heads, each head receives a disjoint slice of the projected representation. Consequently, during training, gradients from each head update only the corresponding slice of the shared projection matrix, resulting in effective per-head specialization despite the shared parameterization.

The L1 weighting parameters $\lambda_h$ are initialized to zero and learned independently for each attention head. Each $\lambda_h$ scales only the L1 penalty computed from that head's slice of the projected query and key vectors. Because the weighted value outputs from all heads are combined and passed to higher Transformer blocks, the geometry of the query and key vectors evolves across layers (see section 5). Allowing $\lambda_h$ to be learned enables each head to adaptively influence the L1 term based on the statistical structure of the representations it receives.

As will be seen below, L1 performs better than RBF-L2, which, by itself, validates the implementation of L1. L1's advantage is that it preserves proportional differences between queries and keys, while L2 either enlarges or reduces those dimensional differences, according to the differences being greater or less than one. Thus, the L2 term over-penalizes large magnitude coordinate differences and under-penalizes small ones.

The wikitext-2-raw-v1 dataset was selected for language modeling. (Merity, et al., 2016) The dataset contains 36718 rows of text for training, 3760 for validation and 4358 for testing. We used the SentencePiece tokenizer of Kudo and Richardson (2018) to prepare the raw text for training.

The models were all trained on a Mac Studio M2 Ultra desktop running Keras/TensorFlow and using Apple's Metal environment to access a GPU. At present, Apple computers are constrained to using only a single GPU, which prevents the use of GPU parallelism. The best performing model (described in the next section) has 1,760,470 learnable parameters.

**5. Modeling Results**

A baseline was first established that used the existing Keras class MultiHeadAttention which implements the Vaswani attention equation in (1), using 64 dimensional query, key and value vectors for each attention head. This was followed by an implementation of equation (5) for the RBF-L2 equation softmax scores as described in Kim et. al. (2021), as well as an L1 baseline model using equation (6) to subtract the L1 norms of the same query and key vectors used to compute the dot product scores, but without including any learned weight parameters $\lambda_h$.

As the first three rows of Table 1 show, the L1 Baseline achieves a significant reduction of perplexity vis-a-vis both the Vaswani and the Kim RBF-L2 models, but at a steep cost in compute time. To address the computational cost, the queries and keys in matrices Q and K were projected to a variety of dimensionalities before computing the L1 norm. All such projection models included the learned $\lambda_h$ weight parameters.

This does nothing to escape the inherent quadratic cost of the Vaswani model of $O(T^2 d_{key})$, where T is the input token size and $d_{key}$ the key dimensionality. Nonetheless, the projections reduce the measurable compute time since the projections reduce the complexity to $O(T^2 d_{proj})$, where $d_{proj}$ is the smaller projection dimensionality.

Thus, Table 1 includes seven additional rows for different projections of the queries and keys, including one projection to the full key dimensionality of 64. There is also a final model that is a hybrid of 8-dimensional L1 projections and L2. The projection models offer two benefits. By projecting the queries and keys from $d_k = 64$ to a lower dimensionality, the computational cost of the L1 term is reduced.

The second benefit relates to the learnable parameters of the projection models. When the queries and keys are projected to lower dimensions, their geometry is necessarily smoothed and a certain amount of information is lost. The projection parameters compensate for this loss by learning to specialize the queries and keys for their new role of capturing useful L1 dissimilarities as described in section 3. These L1-specialized query/key embeddings are analogous to those of Shaw et al. (2018) and Huang et al. (2019) that learn to represent the ordinal differences between queries and keys.

| Model | Mins/ Epoch | Time in Vaswani Epochs | Test Perplexity | Perplexity Reduction from Vaswani |
|---|---|---|---|---|
| **Vaswani** | 66.3 | 1.00 | 187.1 | – |
| **Kim RBF-L2** | **67.6** | **1.02x** | 178.1 | 4.8% |
| **L1 Baseline** | 100.7 | 1.52x | 169.2 | 9.6% |
| **$L1_{64}$** | 105.2 | 1.59x | 182.8 | 2.3% |
| **$L1_{16}$** | 81.3 | 1.23x | 182.3 | 2.6% |
| **$L1_{14}$** | 95.1 | 1.43x | 169.0 | 9.7% |
| **$L1_{12}$** | 90.3 | 1.36x | 190.0 | (-1.5%) |
| **$L1_{10}$** | 87 | 1.31x | 179.5 | 4.3% |
| **$L1_{8}$** | 85.3 | 1.29x | **159.9** | **14.5%** |
| **$L1_{6}$** | 78.2 | 1.18x | 185.9 | 0.6% |
| **$L1_{8}$/RBF** | 74.7 | 1.13x | 173.8 | 7.1% |

Table 1: Perplexity and training time comparisons across a variety of attention equations. L1 projection dimensions are shown as subscripts to model names, e.g. $L1_8$ projects the queries and keys from $d_k$ 64 down to 8. Model $L1_8$/RBF uses 8 dimensional projections in the lower heads and the Kim RBF-L2 decrement in the upper heads with no projections.

Only model $L1_{12}$ failed to attain perplexity improvements over the Vaswani model. At some of the smaller dimensionalities, the specializations result in even lower perplexity than the L1 Baseline which does not include projections.

An ablation study on the dimensionality of the projections revealed an approximate optimal size for the model perplexity. As the table shows, the $L1_8$ model reached the lowest perplexity by far, besting Vaswani with a 14.5% reduction, which is more than three times the reduction of the RBF-L2 kernel. However as the table also shows, the perplexity values of the different projections do not lie on a convex surface. Further, it was discovered that different seeding of the training data impacted which dimensionality obtained the lowest overall perplexity. On one run with different seeding, the $L1_{14}$ model performed best.

Neither the success of the smaller dimensionality nor its indeterminate optimality is surprising. Sanjeev Arora (2016) writes that "A striking finding in empirical work on word embeddings is that there is a sweet spot for the dimensionality of word vectors: neither too small, nor too large." Landauer, Foltz, and Laham (1998) illustrate this with the graph shown here in Figure 1 from a synonym test on different dimensionalities of embeddings created from Latent Semantic Analysis. Yin and Shen (2018) explain the existence of this ‘sweet spot’ mathematically as a tradeoff between bias and variance. The loss on the left is due to high bias from having too few dimensions and on the right from high variance from too many.

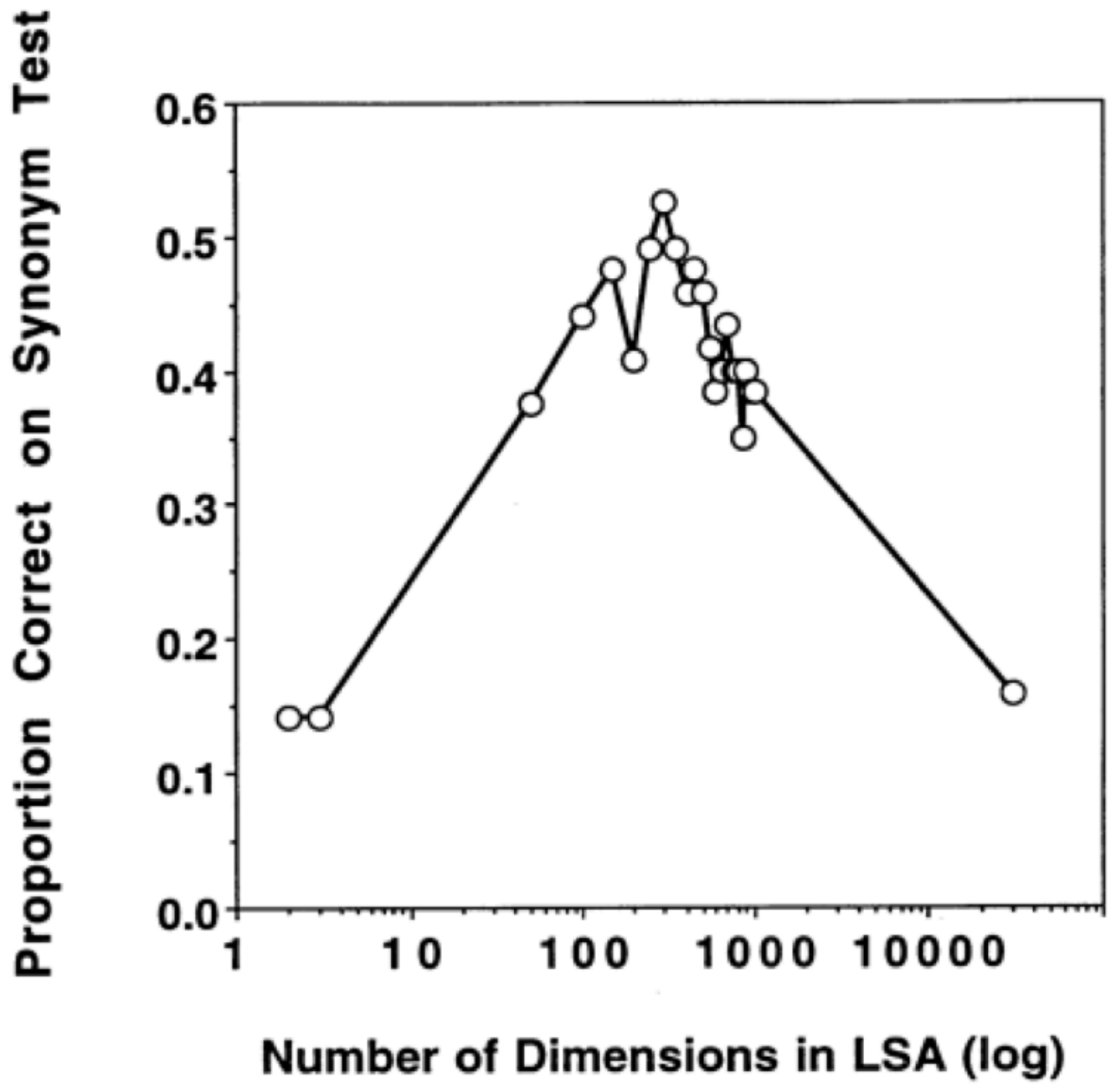

Figure 1: Optimal Dimensionality as a Bias-Variance Tradeoff
From Landauer, Foltz, & Laham (1998)

There are some additional nuances of interest when looking at the results of Table 1. First, while the $L1_{64}$ model trains only slightly slower than the L1 Baseline, it has a much larger perplexity score, though still slightly better than the Vaswani model.

It is nearly the case that the computational cost of the projection models decreases monotonically as the projection dimensionality decreases. But for reasons that are as yet unexplained, the $L1_{16}$ model trains even faster than $L1_8$. This anomaly was reproduced in an additional run. One possible explanation for this counterintuitive performance behavior may be the interaction between our attention implementation and the Apple Matrix Coprocessor (AMX) matrix accelerator, which is optimized for outer-product computations using hardware tiles that operate most efficiently at 16-element widths (Kwon, 2024). However, information on AMX is scarce and this speculation may be difficult to confirm.

As noted above, the learned $\lambda$ parameters are indexed separately for each attention head. These parameters play an important role in the perplexity improvements of the projection models, and when omitted, perplexity suffers. As it turns out, they also stand in an important relationship with the variance of the L1 norms.

Recall that the purpose of the L1 norms is to enhance the quantification of vector similarity by penalizing the dot product values with the weighted norms. The norms make this possible since they capture dimensional differences between vectors. Vectors with large dimensional differences result in large L1 difference norms, and as the query/key vectors converge, so do their norms converge toward zero.

The variance of the norms summarizes the distinctions between tokens, and large norm variance indicates more informative distinctions. As a result, the $\lambda$ values tend to stand in an inverse relationship with the L1 norms. We can see this visually if we plot L1 norm variance vs. $\lambda$ values by epoch, as shown in Figure 2.

The lower block's heads rely more heavily on the L1 norms than the heads in the upper block, as evidenced by the larger $\lambda$ values (right-most vertical scale) in the lower block which are needed to amplify the more subtle Q-K differences. The lower block displays much less norm variance than the upper block. The lower block's smoother curves are more monotonic with more local geometry that is less explosive than the upper block.

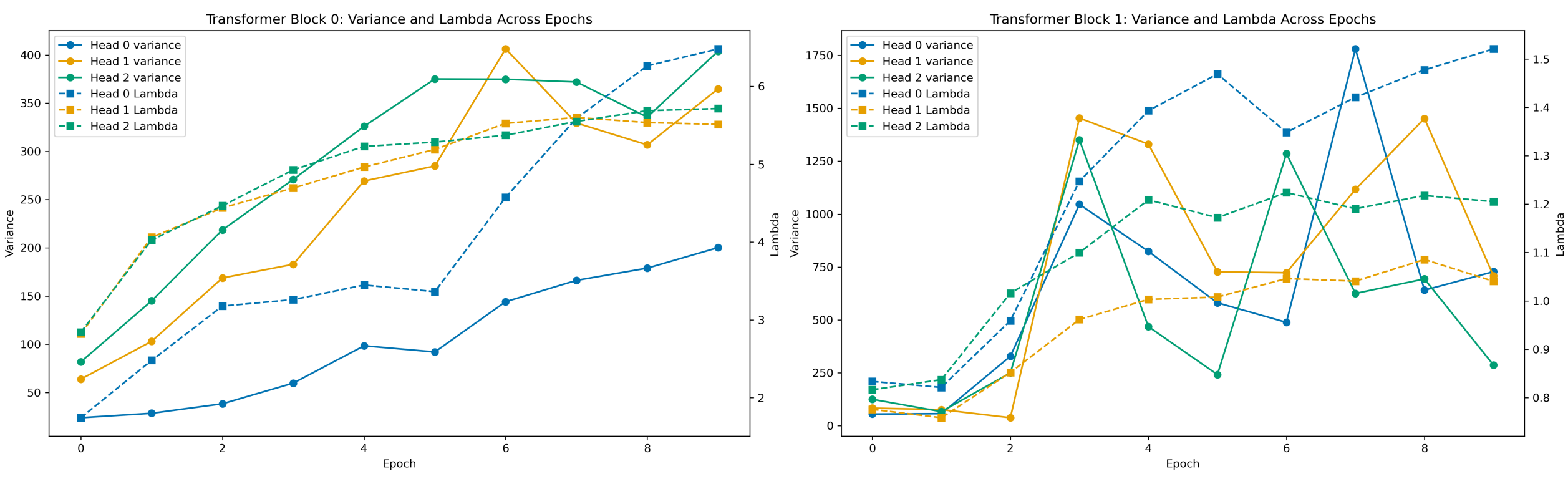

Figure 2a: Lower Block
Norm Variance and Lambda by Epochs

Figure 2b: Upper Block
Norm Variance and Lambda by Epochs

The heads display differing levels of specialization as evidenced by their significantly different geometries. The upper block's heads diverge in behavior from one another much more than in the lower block. This is expected as the upper blocks are specializing on more semantic abstractions, long-range dependencies and global context integration. In contrast, the lower block is specialized more for local, early stage token mixing, short-range dependencies and surface geometric difference relationships. As a result, the lower block with its much larger $\lambda$ values focuses more on the $Q - K$ geometry in the L1 norms than the upper block.

As further evidence that there is a strong interaction between the geometry of the norms and the values of $\lambda$. We computed the Pearson correlation coefficient between per-epoch norm variance and λ values for each head in each block. Table 2 shows the results.

| | Block 0<br>less variance,<br>larger $\lambda$ | Block 1<br>more variance,<br>smaller $\lambda$ |
|---|---|---|
| **Head 0** | 0.9668 | 0.6412 |
| **Head 1** | 0.9232 | 0.7860 |
| **Head 2** | 0.9579 | 0.5079 |

Table 2: Pearson correlations between norm variance and λ values

Here we see extremely large correlations in the lower block, indicating that $\lambda$ closely tracks the L1 geometry, explaining the large $\lambda$ values. The upper block still finds L1 norms useful, but does not track them nearly as much, hence the moderate correlations and smaller $\lambda$ values.

The hybrid $L1_8$/L2 model in Table 1 is of special interest. As we have seen, the lower block heads employ higher $\lambda$ values to focus on the norm differences where they are most useful. The upper block's smaller $\lambda$ values shift the focus to the $Q \cdot K^\top$ term, while still valuing the norms to a lesser extent. However, there is a substantial computational cost to using the L1 norms throughout. Thus, we investigated if there may be a benefit to using the L1 norms for the lower block where the norms are most valued, and shift to the RBF-L2 kernel for the upper block in order to obtain its perplexity benefit over Vaswani without incurring the much larger L1 compute cost.

Table 1 shows that this hybrid approach indeed has the intended effect. Its overall perplexity gain over Vaswani is roughly half the gain from using L1 norms throughout, but the computational overhead is also roughly cut in half. For some applications with limited computational resources, this may be a worthwhile tradeoff.

However, in the absence of such resource constraints, we need not pay the computational penalty when using L1 norms. This is because the L1 term in equation (6) is embarrassingly parallel. The L1 term can be computed in parallel on a separate GPU from the dot product computation, obtaining the much improved perplexity benefit of full L1 norm usage with a negligible compute penalty.

## 6. Limitations

The present study is not the last word on a vector similarity metric. While section 1 presents some limitations of dot product, L1 Augmented Attention is also imperfect. For example, consider these vectors :

$q_4 = \begin{pmatrix} 1 \\ 2 \\ 3 \end{pmatrix}$ $k_6 = \begin{pmatrix} 1 \\ 2 \\ 3 \end{pmatrix}$ $k_7 = \begin{pmatrix} 2 \\ 2 \\ 3 \end{pmatrix}$ The unscaled L1 attention score (before softmax) of the first two is: $q_4 \cdot k_6 - \|q_4 - k_6\|_1 = 14$ while so is $q_4 \cdot k_7 - \|q_4 - k_7\|_1$. And even though their L1 augmented attention scores are equal, no one would claim that $q_4$ is as similar to $k_7$ as it is to $k_6$.

An empirical limitation of this present work is that L1 attention was only tested on a single dataset, wikitext-2-raw-v1, as cited above, and only one modeling type, an autoregressive language model. Further, the architecture of the Transformer model itself is limited and is very small with only 1,760,470 trainable parameters. However, L1 addresses the inductive bias of dot product, which is not dependent on any one dataset or modeling algorithm.

# 7. Conclusions

We have presented clear evidence that a weighted L1 norm decrement to the scaled attention score results in a significant reduction of perplexity in a language model. This reduction comes at a computational cost, which motivated projections of the queries and keys to smaller dimensions in the attention heads. In addition to large reductions in computational cost of the L1 term, the projections also lead to further reductions in perplexity, and the remaining compute penalty can be avoided altogether by leveraging the embarrassingly parallel nature of the L1 penalty term and computing it in parallel with the dot product term.

We noted above how Katharopoulos et al. (2020) present a revision to attention that scales linearly with the input length, rather than quadratically. In their conclusions they write that, "Another line of research to be explored is related to the choice of feature map for linear attention." One of the future extensions of the present work will be to first rework L1 modified attention as a Laplacian kernel with query/key projections using u and v along the lines of $\kappa_{Lap}(q, k) = \exp(-\lambda \|u(q) - v(k)\|_1)$ and combine it with a softmax kernel with the aim of attaining linear scaling with the L1 decrement.

While the evidence presented in this paper only involves a single dataset and a language model, the expectation is that similar benefits will result with additional datasets and modeling types, e.g. text categorization using Transformers. This expectation rests on the simple observation from section 1 that dot product constitutes a flawed similarity metric for vectors, and we pointed out how the L1 norms improve vector similarity matching. Nonetheless, assumptions often lead to errors, and so the next phase of this research will be to gather additional evidence from other datasets as well as other modeling algorithms. We also intend to investigate the possible source of Apple silicon for the anomaly regarding efficiency of computing L1 norms of 16 vs. 8-dimensional vectors. It may be possible to revise the L1 attention code to take advantage of the Apple hardware.